\documentclass[conference]{IEEEtran}
\IEEEoverridecommandlockouts

\usepackage{cite}
\usepackage{amsmath,amssymb,amsfonts}
\usepackage{algorithmic}
\usepackage{graphicx}
\usepackage{textcomp}
\usepackage{xcolor}
\def\BibTeX{{\rm B\kern-.05em{\sc i\kern-.025em b}\kern-.08em
    T\kern-.1667em\lower.7ex\hbox{E}\kern-.125emX}}

\usepackage[utf8]{inputenc} 
\usepackage[T1]{fontenc}    
\usepackage{hyperref}       
\usepackage{url}            
\usepackage{booktabs}       
\usepackage{amsfonts}       
\usepackage{nicefrac}       
\usepackage{microtype}      
\usepackage{xcolor}         
\usepackage{graphicx}
\usepackage{colortbl}
\usepackage{amssymb}
\usepackage{amsmath}
\usepackage{pifont}
\usepackage{caption}

\graphicspath{{figs/}}

\newcommand{\norm}[1]{\left\lVert #1\right\rVert}
\newcommand{\cmark}{\ding{51}}
\newcommand{\xmark}{\ding{55}}

\title{Modular Prompt Learning Improves \\ Vision-Language Models}

\author{\IEEEauthorblockN{Zhenhan Huang}
\IEEEauthorblockA{\textit{Rensselaer Polytechnic Institute}\\
Troy, USA \\
Email: huangz12@rpi.edu}
\and
\IEEEauthorblockN{Tejaswini Pedapati}
\IEEEauthorblockA{\textit{IBM Research}\\
Yorktown Heights, USA \\
Email: tejaswinip@us.ibm.com}
\and
\IEEEauthorblockN{Pin-Yu Chen}
\IEEEauthorblockA{\textit{IBM Research}\\
Yorktown Heights, USA \\
Email: pin-yu.chen@ibm.com}
\and
\IEEEauthorblockN{Jianxi Gao}
\IEEEauthorblockA{\textit{Rensselaer Polytechnic Institute}\\
Troy, USA \\
Email: gaoj8@rpi.edu}
}

\begin{document}

\maketitle
\IEEEpeerreviewmaketitle

\begin{abstract}
    Pre-trained vision-language models are able to interpret visual concepts and language semantics. 
    Prompt learning, a method of constructing prompts for text encoders or image encoders, elicits the potentials of pre-trained models and readily adapts them to new scenarios. Compared to fine-tuning, prompt learning enables the model to achieve comparable or better performance using fewer trainable parameters.
    Besides, prompt learning freezes the pre-trained model and avoids the catastrophic forgetting issue in the fine-tuning. Continuous prompts inserted into the input of every transformer layer (i.e. deep prompts) can improve the performances of pre-trained models on downstream tasks.
    For $i$-th transformer layer, the inserted prompts replace previously inserted prompts in the $(i-1)$-th layer. Although the self-attention mechanism contextualizes newly inserted prompts for the current layer and embeddings from the previous layer's output, removing all inserted prompts from the previous layer inevitably loses information contained in the continuous prompts. In this work, we propose Modular Prompt Learning (MPL) that is designed to promote the preservation of information contained in the inserted prompts. We evaluate the proposed method on base-to-new generalization and cross-dataset tasks. On average of 11 datasets, our method achieves 0.7\% performance gain on the base-to-new generalization task compared to the state-of-the-art method. The largest improvement on the individual dataset is 10.7\% (EuroSAT dataset). Our code is available at \url{https://github.com/Zhenhan-Huang/Modular-Prompt-Learning}.
\end{abstract}

\section{Introduction}

Pre-trained vision-language models (VLMs), such as CLIP \cite{radford2021learning}, ALIGN \cite{jia2021scaling} and FLIP\cite{yao2021filip}, exhibit great potential on various tasks. Owing to the good alignment between visual concepts and language semantics, VLMs are readily transferred to downstream tasks through text prompting \cite{radford2021learning,jia2021scaling,yao2021filip,li2021supervision,singh2022flava} with few shots learning or even zero-shot transfer inference.

Despite the efficacy of manually crafted templates, how to determine the appropriate text prompt to elicit the potential of VLMs is nontrivial. Even a small modification to the text prompt can lead to a pronounced performance variation on the downstream task. For example, modifying the text prompt for the CLIP model from `\texttt{a photo of a [class]}' to `\texttt{a [class]}' remarkably affects the model performance \cite{zhou2022learning}. Continuous prompts (a.k.a. soft prompts) enable an automatic prompting for VLMs. This method has two main advantages: (1) the text embedding is not confined to the embedding of natural language words. It essentially relaxes the search space for the text embedding to be continuous; (2) continuous prompts are not parametrized by pre-trained VLMs and have their own trainable parameters that can be adapted to various downstream tasks. In the natural language processing (NLP) field, continuous prompts have shown significant improvement in the performance of pre-trained models on various downstream tasks \cite{li2021prefix,lester2021power,liu2023gpt,qin2021learning,shin2020autoprompt}. Recently, there is emerging research on applying continuous prompts in vision-language field \cite{zhou2022learning,zhou2022conditional,jia2022visual,khattak2023maple,khattak2023self}.


Deep visual prompt tuning (VPT) \cite{jia2022visual} inserts learnable prompts to each transformer layer's input. Inserted prompts are consistently replaced by newly inserted ones in the following layers. Inputs to $i$-th and $(i+1)$-th transformer layers are:
\begin{equation}
    [\underline{\hspace{2ex}}, \mathbf{I}_{i+1}] = L_{i}([\mathbf{V}_i, \mathbf{I}_i]) \;, [\underline{\hspace{2ex}}, \mathbf{I}_{i+2}] = L_{i+1}([\mathbf{V}_{i+1}, \mathbf{I}_{i+1}]) \;,
\end{equation}
where $\mathbf{I} \in \mathbb{R}^{\xi \times d_v}$ is a collection of image patch embedding, $\mathbf{V} \in \mathbb{R}^{p \times d_v}$ is inserted continuous prompts. The conventional shallow continuous prompt learning method has input to two adjacent transformer layers:
\begin{equation}
    [\mathbf{V}_{i+1}, \mathbf{I}_{i+1}] = L_i([\mathbf{V}_i, \mathbf{I}_i]) \;, [\mathbf{V}_{i+2}, \mathbf{I}_{i+2}] = L_{i+1}([\mathbf{V}_{i+1}, \mathbf{I}_{i+1}]) \;.
\end{equation}
The shallow continuous prompt method inserts prompts into the input of the first transformer layer and those prompts take part in the contextualization process (i.e. self-attention \cite{vaswani2017attention}) of all the following transformer layers. The deep prompt method, on the other hand, consistently replaces continuous prompts with newly inserted ones. A natural question raises \textit{will removing inserted prompts lose information contained it to some extent?} To explore the answer to this question, we propose Modular Prompt Learning (PML) that promotes the preservation of the inserted prompts. There are three main components: adding prompts ($\mathcal{O}_{\rm add}$), carrying prompts ($\mathcal{O}_{\rm cr}$) and removing prompts ($\mathcal{O}_{\rm rm}$). Using a combination of these three components, our proposed method incorporates deep VPT as a special case where the number of inserted prompts by $\mathcal{O}_{\rm add}$ is equal to the number of removed prompts $\mathcal{O}_{\rm rm}$, and the number of prompts by $\mathcal{O}_{\rm cr}$ is always zero. We demonstrate the efficacy of the proposed method on typical representative benchmarks including base-to-new generalization and cross-dataset evaluation. Despite the simplicity of the proposed method, it achieves an average improvement of 0.7\% on 10 datasets in the base-to-new generalization benchmark. The highest performance gain is 10.7\% on the EuroSAT dataset. On the cross-dataset evaluation benchmark, our method is comparable with the state-of-the-art method but with much higher efficiency.

\section{Related Work}
\label{sec:work_relate}

\paragraph{VLM Adaption} VLMs jointly pretrain image-language representations in an unsupervised manner that enables the utilization of an immense amount of data. For example, CLIP \cite{radford2021learning} and ALIGN \cite{jia2021scaling} use \textasciitilde400M and \textasciitilde1B image-text pairs in the pretraining. VLMs exhibit good generalization behavior \cite{radford2021learning,jia2021scaling,yao2021filip,li2021supervision,singh2022flava}. The following researches explore the potential of pre-trained VLMs in challenging downstream tasks such as semantic segmentation \cite{rao2022denseclip} visual question answering \cite{tan2019lxmert}, image retrieval \cite{lu2019vilbert}, and visual grounding \cite{yao2024cpt}.

\paragraph{Prompt Learning} In the vision-language field, the main focus is on continuous prompts. The pioneering work for the text prompt is CoOp \cite{zhou2022learning} which replaces embeddings of the CLIP text template with continuous prompts. The pioneering work for the visual prompt is VPT which prepends continuous prompts to image patch embedding. There are emerging researches trying to boost the performance of continuous prompts. Based on CoOp work, CoCoOp \cite{zhou2022conditional} adds special tokens conditioned on image input. MaPLe \cite{khattak2023maple} extends the coupling effect between inserted continuous prompts for language branch and that for visual branch to more transformer layers by using deep prompts. ProGrad \cite{zhu2023prompt} and PromptSRC \cite{khattak2023self} use knowledge distillation (KD) that has pre-trained CLIP as teacher's model.

\section{Revisiting CLIP}
\label{sec:clip_revisit}

\paragraph{Zero-Shot CLIP} The CLIP model \cite{radford2021learning} is comprised of an image encoder and a text encoder. Given a labeled image dataset $\mathcal{D} = \{\mathbf{x}_i, y_i\}^{N}_{i=1}$ that has $C$ classes in total $\mathcal{C} = \{c_j\}^C_{j=1}$. In the zero-shot transfer inference on downstream tasks, templates such as \texttt{a photo of a [class]} are used to prompt the text encoder. A text prompt $t_j$ is fed into the text encoder. The image encoder takes an image $\mathbf{x} \in \mathbb{R}^{H \times W \times 3}$ as the input. The text embedding and image embedding are linearly transformed to get the joint multimodal embedding. We denote the text encoder and the following linear transformation as $g(\cdot)$, the image encoder and the following linear transformation as $f(\cdot)$. The normalized image embedding is denoted by $\mathbf{u} = f(\mathbf{x}) / \norm{f(\mathbf{x})}_2 \in \mathbb{R}^{d}$ while the normalized text embedding denoted by $\mathbf{w}_j = g(t_j) / \norm{g(t_j)}_2 \in \mathbb{R}^{d}$. The probability prediction $p(\hat{y}|\mathbf{x})$ is based on the pairwise cosine similarity between the text embedding of all possible classes $\mathbf{W} = [\mathbf{w}_1, \ldots, \mathbf{w}_{C}] \in \mathbb{R}^{C \times d}$ and the image embedding:
\begin{equation}\label{eq:clip_predict}
p(\hat{y} | \mathbf{x}) = \frac{\exp(\mathbf{u}\mathbf{w}^T_{\hat{y}} / \tau)}{\sum_{j=1}^{C}\exp(\mathbf{u} \mathbf{w}^T_j / \tau)} \;,
\end{equation}
where $\tau$ is the temperature parameter.

\paragraph{Continuous Prompts for CLIP} Continuous prompts for the text encoder are generally initialized using discrete prompt tokens $[W_i]$. For the text encoder of the CLIP model, the commonly used text prompt $t$ \cite{zhou2022learning,zhou2022conditional,khattak2023self,zhu2023prompt,miyai2024locoop,ma2024swapprompt}  used for the embedding is:
\begin{equation}
    t = \{[W_1], [W_2], \ldots, [W_M], [\text{class}]\} \;.
\end{equation}
Here $M$ is the number of discrete prompt tokens. The text prompt containing class name tokens and discrete prompt tokens is mapped into input embeddings for the text encoder:
\begin{equation}
    [\mathbf{e}(W_1), \mathbf{e}(W_2), \ldots, \mathbf{e}(W_M), \mathbf{e}(\text{class})] \in \mathbb{R}^{(M+1) \times d_t} \;,
\end{equation}
where $[\cdot, \cdot]$ denotes the concatenation operation and $d_t$ is the feature dimension for the text encoder. $\mathbf{e}(W_i)$ is the embedding of the word token $W_i$.

Continuous prompts for the image encoder are randomly initialized and inserted into image patch embeddings \cite{jia2022visual}. We denote a collection of prompts as $\mathbf{V} = \{\mathbf{v}_i \mid i \in \mathbb{N}, 1 \le i \le p\}$ and a collection of image patch embedding as $\mathbf{I} = \{\mathbf{i}_j \mid \in \mathbb{N}, 1 \le j \le \xi\}$. The input to the image encoder of the CLIP model is:
\begin{equation}
    [\mathbf{V}, \mathbf{I}] \in \mathbb{R}^{(p+\xi) \times d_v} \;,
\end{equation}
where $d_v$ is the feature dimension for the image encoder.

\section{Modular Prompt Learning}
\label{sec:modular}

\begin{figure*}[thb]
    \centering
    \includegraphics[width=0.8\linewidth]{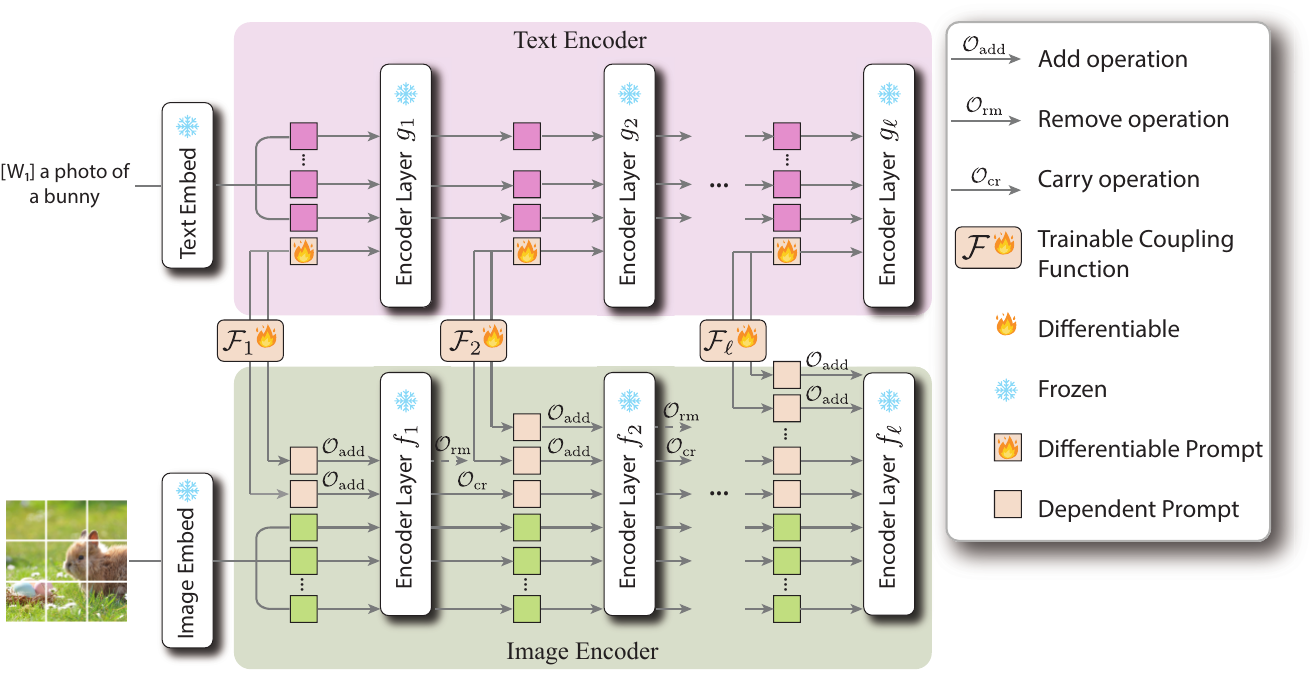}
    \caption{Illustration of the proposed framework. We use deep prompts with coupling function $\mathcal{F}$. The coupling function bridges prompts for the language branch to those for the visual branch. Three operations $\mathcal{O}_{\rm add}$, $\mathcal{O}_{\rm rm}$ and $\mathcal{O}_{\rm cr}$ are applied to enable the varying number of continuous prompts inserted to transformer layers. $\mathcal{O}_{\rm add}$ is applied to inputs of transformer layers while $\mathcal{O}_{\rm rm}$ and $\mathcal{O}_{\rm cr}$ is applied to the output of transformer layers.}
    \label{fig:ct_prompt}
\end{figure*}

The classic way of applying deep visual continuous prompts is firstly proposed by VPT \cite{jia2022visual}, where inserted continuous prompts for $i$-th transformer layer are replaced by those for $(i+1)$-th transformer layer. We postulate that removing inserted prompts and relying on multi-head self-attention (MSA) \cite{vaswani2017attention} is not enough to preserve the information contained in the continuous prompts. To test the proposed postulation, we proposed three operations: adding operation $\mathcal{O}_{\rm add}$, removing operation $\mathcal{O}_{\rm rm}$ and carrying forward operation $\mathcal{O}_{\rm cr}$. The motivation for applying three operations is to enable models to preserve inserted prompts, thereby keeping the information contained within them.

\paragraph{Adding Operation $\mathcal{O}_{\rm add}$} $\mathcal{O}_{\rm add}$ inserts continuous prompts to the transformer layers' inputs. We denote the $i$-th transformer layer as $L_i(\cdot)$, $\mathcal{O}_{\rm add}$ is formulated as $[\mathbf{V}_i, \mathbf{I}_i]$ given the input to the opeartion $\mathbf{I}_i$, where $\mathbf{V}_i \in \mathbb{R}^{p_i \times d_v}$ and $\mathbf{I}_i \in \mathbb{R}^{\xi_i \times d_v}$. $\mathcal{O}_{\rm add}$ increase the context length from $\xi_i$ to $\xi_i + p_i$. $\mathbf{V}_i$ is differentiable (i.e. trainable) and affects $\mathbf{I}_{i+1}$ by MSA.

\paragraph{Removing Operation $\mathcal{O}_{\rm rm}$} $\mathcal{O}_{\rm rm}$ removes the inserted prompts and is formulated as:
\begin{equation}
    \begin{split}
    [\underline{\hspace{2ex}}, \mathbf{V}^{\prime}_{i+1}, \mathbf{I}_{i+1}] = L_i([\mathbf{V}_i, \mathbf{I}_i]) \;, \\ \mathbf{V}^{\prime} \in \mathbb{R}^{p^{\prime}_{i+1} \times d_v}, \; \quad 0 \le p^{\prime}_{i+1} \le p_i, \\
    \end{split}
\end{equation}
where $\mathbf{V}^{\prime}_{i+1}$ is the remaining continuous prompts whose number is equal to the number of inserted prompts by $\mathcal{O}_{\rm add}$ minus the number of removed prompts by $\mathcal{O}_{\rm rm}$. The input to the operation is $L_i([\mathbf{V}_i, \mathbf{I}_i])$ and the output is $[\mathbf{V}^{\prime}_{i+1}, \mathbf{I}_{i+1}]$. When $p^{\prime}_{i+1} = 0$, it is the same as the traditional deep visual prompt. When $p^{\prime}_{i+1} = p_i$ and the prompt depth is 1, it becomes the traditional shallow visual prompt.

\paragraph{Carrying Operation $\mathcal{O}_{\rm cr}$} $\mathcal{O}_{\rm add}$, $\mathcal{O}_{\rm rm}$ and $\mathcal{O}_{\rm cr}$ are not mutually independent. $\mathcal{O}_{\rm cr}$ depends on $\mathcal{O}_{\rm add}$ and $\mathcal{O}_{\rm rm}$. It is formulated as $[\mathbf{V}^{\prime}_{i+1}, \mathbf{I}_{i+1}]$ given the input to the operation $[\mathbf{V}^{\prime}_{i+1}, \mathbf{I}_{i+1}]$ . $\mathcal{O}_{\rm cr}$ essentially leads to possibly different number of continuous prompts for different layers.

Three operations are summarized in the Table \ref{tab:ops}. The row of MSA indicates if the newly inserted continuous prompts take part in the MSA of the following transformer layers in addition to the current transformer layer. $\mathcal{O}_{\rm add}$ is related to the input of the transformer layer while $\mathcal{O}_{\rm rm}$ and $\mathcal{O}_{\rm cr}$ are associated with the output. Hence, only $\mathcal{O}_{\rm rm}$ and $\mathcal{O}_{\rm cr}$ determine if the inserted prompts are used in the MSA of following transformer layers.

Figure \ref{fig:ct_prompt} shows the proposed framework. Inspired by existing works \cite{zang2022unified,khattak2023maple}, we use multi-modal continuous prompts to enhance the coupling effect between the prompts for text encoder and those for image encoder. Three operations are applied to each transformer layer in the visual branch. $\mathcal{O}_{\rm add}$ inserts continuous prompts and the context length increases. The inserted prompts can be removed by $\mathcal{O}_{\rm rm}$. $\mathcal{O}_{\rm cr}$ carries forward the prompts to the next layer. Inserted prompts by $\mathcal{O}_{\rm add}$ are associated with those for the language branch by coupling function. We denote the coupling function for $i$-th layer of $\ell$ transformer layers as $\mathcal{F}_i: \mathbb{R}^{d_t} \rightarrow \mathbb{R}^{d_v}$, where $i \in \mathbb{N}, 1 \le i \le \ell$.

\begin{table}[thb]
    \centering
    \caption{Three operations used in the proposed framework. $\mathcal{O}_{\rm add}$ and $\mathcal{O}_{\rm rm}$ changes the number of embeddings while $\mathcal{O}_{\rm cr}$ keep the number of embedding constant. $\mathcal{O}_{\rm add}$ and $\mathcal{O}_{\rm cr}$ enable continuous prompts to participate the MSA in the following transformer layer while $\mathcal{O}_{\rm rm}$ prevents the prompts from MSA.}
    \label{tab:ops}
    \resizebox{.7\linewidth}{!}{%
    \begin{tabular}{cccc}
        \hline
        Operation & $\mathcal{O}_{\rm add}$ & $\mathcal{O}_{\rm rm}$ & $\mathcal{O}_{\rm cr}$ \\
        \hline
        Input     & $[\mathbf{I}]$ & $[\mathbf{V}, \mathbf{I}]$ & $[\mathbf{V}^{\prime}, \mathbf{I}]$\\
        Output    & $[\mathbf{V}, \mathbf{I}]$ & $[\mathbf{V}^{\prime}, \mathbf{I}]$ & $[\mathbf{V}^{\prime}, \mathbf{I}]$\\
        MSA       & \cmark or \xmark & \xmark & \cmark \\
        \hline
    \end{tabular}
    }
\end{table}


\begin{table*}[thb]
    \centering
    \caption{Test accuracy on the best-to-novel generalization task. Results are averaged by 6 runs. \colorbox{green!10}{Green color} denotes the best performance. \colorbox{blue!10}{Blue color} denotes the second best performance.}
    \label{tab:base_to_new}
    \resizebox{\linewidth}{!}{%
        \begin{tabular}{ccccccccccccc}
            \toprule
                       & Caltech101 & DTD & EuroSAT & Aircraft & Food & Flowers & Pets & Cars & Sun & UCF & ImageNet & Avg \\
            \midrule
             CoOp      & 89.98 & 42.65 & 48.73 & 22.37 & 83.82 & 58.27 & 92.45 & 55.28 & 62.44 & 54.12 & 59.72 & 60.89 \\
             CoCoOp    & 93.53 & 53.93 & 63.27 & \cellcolor{blue!10} 33.38 & 91.40 & 71.65 & \cellcolor{green!10} 97.78 & 73.35 & 76.63 & 73.27 & 70.43 & 72.60 \\
             MaPLe     & \cellcolor{green!10} 94.08 & \cellcolor{blue!10} 59.72 & 71.98 & 30.88 & \cellcolor{green!10} 91.72 & 73.52 & 97.65 & 73.65 & 78.38 & 77.92 & \cellcolor{blue!10} 70.77 & 74.57 \\
             PromptSRC & 93.92 & \cellcolor{green!10} 61.82 & \cellcolor{blue!10} 72.05 & 29.03 & 91.52 & \cellcolor{green!10} 76.57 & 97.28 & \cellcolor{green!10} 75.03 & \cellcolor{blue!10} 78.48 & \cellcolor{blue!10} 78.28 & 70.45 & \cellcolor{blue!10}74.95 \\
             Ours      & \cellcolor{blue!10} 94.20 & 58.13 & \cellcolor{green!10} 79.73 & \cellcolor{green!10} 34.80 & \cellcolor{blue!10} 91.70 & \cellcolor{blue!10} 74.10 & \cellcolor{blue!10} 97.77 & \cellcolor{blue!10}73.72 & \cellcolor{green!10} 78.52 & \cellcolor{green!10} 78.35 & \cellcolor{green!10} 70.92 &  \cellcolor{green!10} 75.63 \\
            \bottomrule
        \end{tabular}
    }
\end{table*}

\begin{table*}[thb]
    \centering
    \caption{Test accuracy on the cross-dataset task. Results are averaged by 3 runs. \colorbox{green!10}{Green color} denotes the best performance. \colorbox{blue!10}{Blue color} denotes the second best performance.}
    \label{tab:cross_dataset}
    \resizebox{.94\linewidth}{!}{%
    \begin{tabular}{cccccccccccc}
        \toprule
                   & Caltech101 & DTD & EuroSAT & Aircraft & Food & Flowers & Pets & Cars & Sun & UCF & Avg \\
        \midrule
         CoOp      & 89.47 & 37.00 & 41.07 & 14.08 & 80.97 & 58.60 & 85.87 & 59.33 & 57.93 & 60.00 & 58.43 \\
         CoCoOp    & \cellcolor{blue!10} 94.03 & 45.42 & 44.70 & 22.97 & 86.13 & \cellcolor{blue!10} 71.93 & \cellcolor{green!10} 90.70 & \cellcolor{blue!10} 65.57 & 64.93 & 67.57 & 65.40 \\
         MaPLe     & 93.50 & 44.93 & 44.80 & \cellcolor{blue!10} 23.74 & 85.70 & 71.63 & 90.24 & 64.90 & 67.17 & 68.27 & 65.49 \\
         PromptSRC &  93.66 & \cellcolor{green!10} 46.17 & \cellcolor{blue!10} 46.40 & \cellcolor{green!10} 24.50 & \cellcolor{green!10} 86.23 & 70.50 & 90.20 & 65.37 & \cellcolor{green!10} 67.50 & \cellcolor{green!10} 68.83 & \cellcolor{blue!10} 65.94 \\
         Ours      & \cellcolor{green!10} 94.10 & \cellcolor{blue!10} 45.47 & \cellcolor{green!10} 49.00 & 23.43 & \cellcolor{blue!10} 86.20 & \cellcolor{green!10} 72.20 & \cellcolor{blue!10}90.33 & \cellcolor{green!10} 65.93 & \cellcolor{blue!10} 67.43 & \cellcolor{blue!10} 68.40 & \cellcolor{green!10} 66.25 \\
        \bottomrule
    \end{tabular}
    }
\end{table*}

\section{Experiments}

\paragraph{Datasets} We use 11 publicly available labeled image datasets: ImageNet \cite{deng2009imagenet}, Caltech101 \cite{fei2004learning}, OxfordPets \cite{parkhi2012cats}, StandfordCars \cite{krause20133d}, Flowers102 \cite{nilsback2008automated}, Food101 \cite{bossard2014food}, FGVCAircraft \cite{maji2013fine}, SUN397 \cite{xiao2010sun}, DTD \cite{cimpoi2014describing}, EuroSAT \cite{helber2019eurosat} and UCF101 \cite{soomro2012ucf101}. These datasets cover generic objects (ImageNet and Caltech101), fine-grained categorization (OxfordPets, StandfordCars, Flowers101, Food101 and FGVCAircraft), scene recognition (SUN397), action recognition (UCF101), and specialized tasks including textures (DTD) as well as satellite imagery (EuroSAT).

\paragraph{Implementation Details} We use few-shot learning in all experiments. Training data are randomly sampled and test accuracy is evaluated on the original full test dataset. Following baseline methods, we use 16 shots and a pre-trained ViT-B/16 CLIP model where $d_t = 512$, $d_v = 768$ and $d = 512$. The text prompt is \texttt{[$W_1$] a photo of a [class]} (i.e. $M = 1$). The number of inserted prompts for each transformer layer of the text encoder is 1 and that for the image encoder is 2. We remove 1 continuous prompt for the output of each transformer layer. The used coupling function $\mathcal{F}$ is the affine function. The prompt depth is $d_{\rm prompt} = 9$. We use SGD optimizer \cite{loshchilov2016sgdr} with an initial warm-up stage, cosine learning rate scheduler and a learning rate of $3.5 \times 10^{-3}$. The batch size for the training dataset is 4 and that for the test dataset is 100. In the base-to-new generalization task, we train the model for 5 epochs and the number of epochs is 2 for cross-dataset evaluation task. The computational platform has AMD EPYC 7232P CPU and a single NVIDIA A40 GPU.

\paragraph{Baseline Methods} We compare our methods with CoOp \cite{zhou2022learning}, CoCoOp \cite{zhou2022conditional}, MaPLe \cite{khattak2023maple} and PromptSRC \cite{khattak2023self}. CoOp replaces the entire text template with continuous prompts. CoCoOp introduces additional terms in the continuous prompt calculation that conditions on the input image. MaPLe bridges the continuous prompts for text encoder to those for image encoder. PromptSRC uses the pre-trained CLIP model and apply KD to regularize inserted prompts.

\subsection{Base-to-New Generalization}

For each dataset, the base-to-new generalization splits the classes equally into two groups: one group is considered as base class for training while the other group is considered as new class for testing. The base-to-new generalization task examines the generalization ability of a model on a specific dataset.

We notice that in the base-to-new generalization task, the model performance tends to have a large variation. We report the performance over 6 runs. Table \ref{tab:base_to_new} shows the test accuracy for new classes using our proposed method in comparison with baseline methods. On 10 out of 11 datasets, our method ranks within top 2. The highest improvement compared to the SOTA method on the individual dataset EuroSAT is 10.7\%. The average test accuracy, compared to the SOTA, increases from 74.95\% to 75.63\%.

Figure \ref{fig:perf_compare} \textit{left} shows the average test accuracy with respect to the average running time. Transformer models have a quadratic complexity. CoOp and CoCoOp insert 16 continuous prompts, which result in the longest running time. Overall, our method achieves high efficiency and good efficacy.

\subsection{Cross-Dataset Evaluation}

In cross-dataset transfer, models are trained on ImageNet in a supervised fashion and directly evaluated on other datasets without any dataset-specific fine-tuning. Compared to the base-to-new generalization, the cross-dataset transfer has a remarkably larger distribution shift between the training dataset and test dataset. Hence, this taks is much more challenging.

The average test accuracy drops for all methods compared to base-to-new generalization methods as shown in Table \ref{tab:cross_dataset}. Our proposed method can achieve the highest average test accuracy of 66.25\%. On EuroSAT dataset, our method improves the SOTA method by 5.6\%. Similar to the base-to-new generalization task, recent works have a similar performance on Caltech101, Food101 and OxfordPets. On EuroSAT dataset, introducing continuous prompt has a pronounced effect on the model performance. Our proposed method also shows a strong strength on this dataset in the cross-dataset evaluation task.

Figure \ref{fig:perf_compare} \textit{right} shows the test accuracy associated with the running time. Similar as the base-to-novel generalization task, our method shows good performance with short running time.

\begin{minipage}{\linewidth}
    \captionsetup{type=figure}
    \centering
    \begin{minipage}{.405\linewidth}
        \centering
        \includegraphics[width=\linewidth]{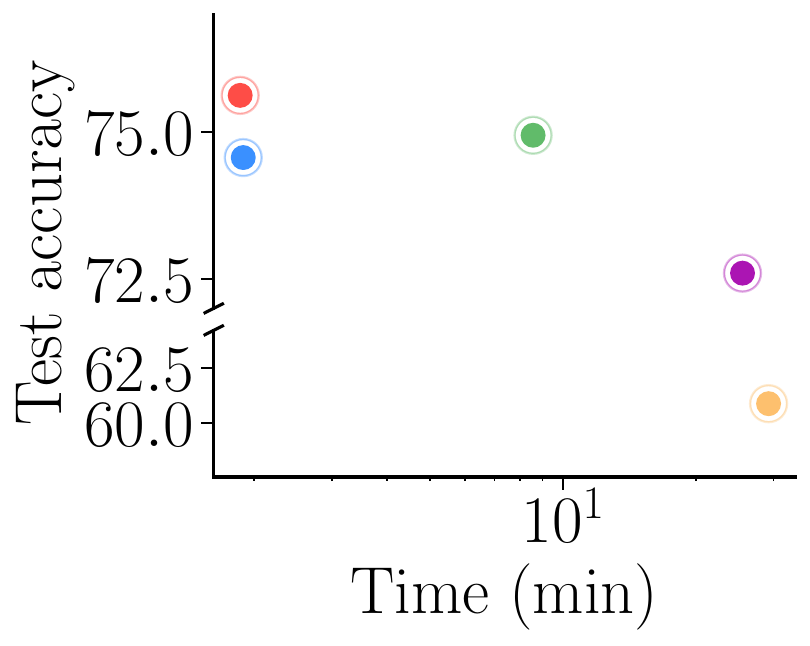}
    \end{minipage}%
    \begin{minipage}{.595\linewidth}
        \centering
        \includegraphics[width=\linewidth]{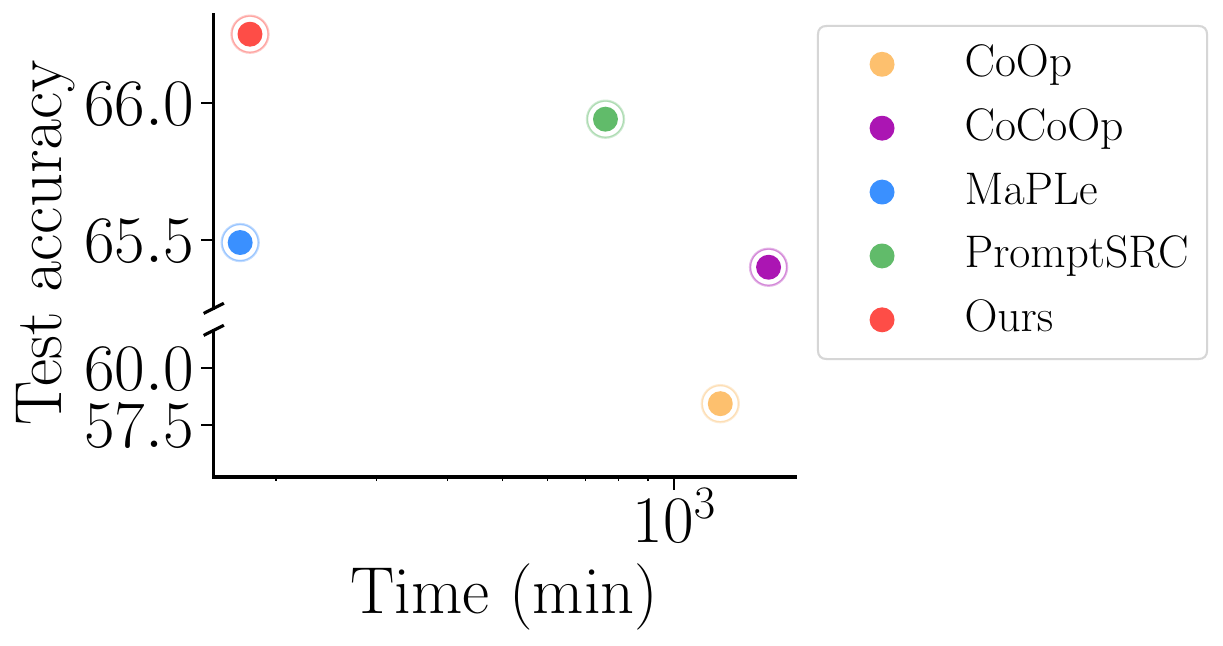}
    \end{minipage}
    \captionof{figure}{Performance on base-to-new generalization (\textit{left}) and cross-dataset evaluation (\textit{right}) with respect to the average time for training. For base-to-new generalization task, the running time is averaged over 10 datasets. For cross-dataset evaluation task, the running time is the time for training model on ImageNet dataset.}
    \label{fig:perf_compare}
\end{minipage}

\section{Conclusion}

We propose a method dubbed modular prompt learning that consists of three operations $\mathcal{O}_{\rm add}$, $\mathcal{O}_{\rm rm}$ and $\mathcal{O}_{\rm cr}$. By introducing these three operations, the number of inserted prompts can very from one transformer layer to another. The results of the base-to-new generalization task and cross-dataset evaluation task exhibit the advantages of the proposed method. The classic way of consistently adding the same number of continuous prompts as removed ones might not lead to the optimal performance.


\bibliography{ref.bib}

\begin{thebibliography}{10}

\bibitem{radford2021learning}
Alec Radford, Jong~Wook Kim, Chris Hallacy, Aditya Ramesh, Gabriel Goh,
  Sandhini Agarwal, Girish Sastry, Amanda Askell, Pamela Mishkin, Jack Clark,
  et~al.
\newblock Learning transferable visual models from natural language
  supervision.
\newblock In {\em International conference on machine learning}, pages
  8748--8763. PMLR, 2021.

\bibitem{jia2021scaling}
Chao Jia, Yinfei Yang, Ye~Xia, Yi-Ting Chen, Zarana Parekh, Hieu Pham, Quoc Le,
  Yun-Hsuan Sung, Zhen Li, and Tom Duerig.
\newblock Scaling up visual and vision-language representation learning with
  noisy text supervision.
\newblock In {\em International conference on machine learning}, pages
  4904--4916. PMLR, 2021.

\bibitem{yao2021filip}
Lewei Yao, Runhui Huang, Lu~Hou, Guansong Lu, Minzhe Niu, Hang Xu, Xiaodan
  Liang, Zhenguo Li, Xin Jiang, and Chunjing Xu.
\newblock Filip: Fine-grained interactive language-image pre-training.
\newblock {\em arXiv preprint arXiv:2111.07783}, 2021.

\bibitem{li2021supervision}
Yangguang Li, Feng Liang, Lichen Zhao, Yufeng Cui, Wanli Ouyang, Jing Shao,
  Fengwei Yu, and Junjie Yan.
\newblock Supervision exists everywhere: A data efficient contrastive
  language-image pre-training paradigm.
\newblock {\em arXiv preprint arXiv:2110.05208}, 2021.

\bibitem{singh2022flava}
Amanpreet Singh, Ronghang Hu, Vedanuj Goswami, Guillaume Couairon, Wojciech
  Galuba, Marcus Rohrbach, and Douwe Kiela.
\newblock Flava: A foundational language and vision alignment model.
\newblock In {\em Proceedings of the IEEE/CVF Conference on Computer Vision and
  Pattern Recognition}, pages 15638--15650, 2022.

\bibitem{zhou2022learning}
Kaiyang Zhou, Jingkang Yang, Chen~Change Loy, and Ziwei Liu.
\newblock Learning to prompt for vision-language models.
\newblock {\em International Journal of Computer Vision}, 130(9):2337--2348,
  2022.

\bibitem{li2021prefix}
Xiang~Lisa Li and Percy Liang.
\newblock Prefix-tuning: Optimizing continuous prompts for generation.
\newblock {\em arXiv preprint arXiv:2101.00190}, 2021.

\bibitem{lester2021power}
Brian Lester, Rami Al-Rfou, and Noah Constant.
\newblock The power of scale for parameter-efficient prompt tuning.
\newblock {\em arXiv preprint arXiv:2104.08691}, 2021.

\bibitem{liu2023gpt}
Xiao Liu, Yanan Zheng, Zhengxiao Du, Ming Ding, Yujie Qian, Zhilin Yang, and
  Jie Tang.
\newblock Gpt understands, too.
\newblock {\em AI Open}, 2023.

\bibitem{qin2021learning}
Guanghui Qin and Jason Eisner.
\newblock Learning how to ask: Querying lms with mixtures of soft prompts.
\newblock {\em arXiv preprint arXiv:2104.06599}, 2021.

\bibitem{shin2020autoprompt}
Taylor Shin, Yasaman Razeghi, Robert~L Logan~IV, Eric Wallace, and Sameer
  Singh.
\newblock Autoprompt: Eliciting knowledge from language models with
  automatically generated prompts.
\newblock {\em arXiv preprint arXiv:2010.15980}, 2020.

\bibitem{zhou2022conditional}
Kaiyang Zhou, Jingkang Yang, Chen~Change Loy, and Ziwei Liu.
\newblock Conditional prompt learning for vision-language models.
\newblock In {\em Proceedings of the IEEE/CVF conference on computer vision and
  pattern recognition}, pages 16816--16825, 2022.

\bibitem{jia2022visual}
Menglin Jia, Luming Tang, Bor-Chun Chen, Claire Cardie, Serge Belongie, Bharath
  Hariharan, and Ser-Nam Lim.
\newblock Visual prompt tuning.
\newblock In {\em European Conference on Computer Vision}, pages 709--727.
  Springer, 2022.

\bibitem{khattak2023maple}
Muhammad~Uzair Khattak, Hanoona Rasheed, Muhammad Maaz, Salman Khan, and
  Fahad~Shahbaz Khan.
\newblock Maple: Multi-modal prompt learning.
\newblock In {\em Proceedings of the IEEE/CVF Conference on Computer Vision and
  Pattern Recognition}, pages 19113--19122, 2023.

\bibitem{khattak2023self}
Muhammad~Uzair Khattak, Syed~Talal Wasim, Muzammal Naseer, Salman Khan,
  Ming-Hsuan Yang, and Fahad~Shahbaz Khan.
\newblock Self-regulating prompts: Foundational model adaptation without
  forgetting.
\newblock In {\em Proceedings of the IEEE/CVF International Conference on
  Computer Vision}, pages 15190--15200, 2023.

\bibitem{vaswani2017attention}
Ashish Vaswani, Noam Shazeer, Niki Parmar, Jakob Uszkoreit, Llion Jones,
  Aidan~N Gomez, {\L}ukasz Kaiser, and Illia Polosukhin.
\newblock Attention is all you need.
\newblock {\em Advances in neural information processing systems}, 30, 2017.

\bibitem{rao2022denseclip}
Yongming Rao, Wenliang Zhao, Guangyi Chen, Yansong Tang, Zheng Zhu, Guan Huang,
  Jie Zhou, and Jiwen Lu.
\newblock Denseclip: Language-guided dense prediction with context-aware
  prompting.
\newblock In {\em Proceedings of the IEEE/CVF conference on computer vision and
  pattern recognition}, pages 18082--18091, 2022.

\bibitem{tan2019lxmert}
Hao Tan and Mohit Bansal.
\newblock Lxmert: Learning cross-modality encoder representations from
  transformers.
\newblock {\em arXiv preprint arXiv:1908.07490}, 2019.

\bibitem{lu2019vilbert}
Jiasen Lu, Dhruv Batra, Devi Parikh, and Stefan Lee.
\newblock Vilbert: Pretraining task-agnostic visiolinguistic representations
  for vision-and-language tasks.
\newblock {\em Advances in neural information processing systems}, 32, 2019.

\bibitem{yao2024cpt}
Yuan Yao, Ao~Zhang, Zhengyan Zhang, Zhiyuan Liu, Tat-Seng Chua, and Maosong
  Sun.
\newblock Cpt: Colorful prompt tuning for pre-trained vision-language models.
\newblock {\em AI Open}, 5:30--38, 2024.

\bibitem{zhu2023prompt}
Beier Zhu, Yulei Niu, Yucheng Han, Yue Wu, and Hanwang Zhang.
\newblock Prompt-aligned gradient for prompt tuning.
\newblock In {\em Proceedings of the IEEE/CVF International Conference on
  Computer Vision}, pages 15659--15669, 2023.

\bibitem{miyai2024locoop}
Atsuyuki Miyai, Qing Yu, Go~Irie, and Kiyoharu Aizawa.
\newblock Locoop: Few-shot out-of-distribution detection via prompt learning.
\newblock {\em Advances in Neural Information Processing Systems}, 36, 2024.

\bibitem{ma2024swapprompt}
Xiaosong Ma, Jie Zhang, Song Guo, and Wenchao Xu.
\newblock Swapprompt: Test-time prompt adaptation for vision-language models.
\newblock {\em Advances in Neural Information Processing Systems}, 36, 2024.

\bibitem{zang2022unified}
Yuhang Zang, Wei Li, Kaiyang Zhou, Chen Huang, and Chen~Change Loy.
\newblock Unified vision and language prompt learning.
\newblock {\em arXiv preprint arXiv:2210.07225}, 2022.

\bibitem{deng2009imagenet}
Jia Deng, Wei Dong, Richard Socher, Li-Jia Li, Kai Li, and Li~Fei-Fei.
\newblock Imagenet: A large-scale hierarchical image database.
\newblock In {\em 2009 IEEE conference on computer vision and pattern
  recognition}, pages 248--255. Ieee, 2009.

\bibitem{fei2004learning}
Li~Fei-Fei, Rob Fergus, and Pietro Perona.
\newblock Learning generative visual models from few training examples: An
  incremental bayesian approach tested on 101 object categories.
\newblock In {\em 2004 conference on computer vision and pattern recognition
  workshop}, pages 178--178. IEEE, 2004.

\bibitem{parkhi2012cats}
Omkar~M Parkhi, Andrea Vedaldi, Andrew Zisserman, and CV~Jawahar.
\newblock Cats and dogs.
\newblock In {\em 2012 IEEE conference on computer vision and pattern
  recognition}, pages 3498--3505. IEEE, 2012.

\bibitem{krause20133d}
Jonathan Krause, Michael Stark, Jia Deng, and Li~Fei-Fei.
\newblock 3d object representations for fine-grained categorization.
\newblock In {\em Proceedings of the IEEE international conference on computer
  vision workshops}, pages 554--561, 2013.

\bibitem{nilsback2008automated}
Maria-Elena Nilsback and Andrew Zisserman.
\newblock Automated flower classification over a large number of classes.
\newblock In {\em 2008 Sixth Indian conference on computer vision, graphics \&
  image processing}, pages 722--729. IEEE, 2008.

\bibitem{bossard2014food}
Lukas Bossard, Matthieu Guillaumin, and Luc Van~Gool.
\newblock Food-101--mining discriminative components with random forests.
\newblock In {\em Computer Vision--ECCV 2014: 13th European Conference, Zurich,
  Switzerland, September 6-12, 2014, Proceedings, Part VI 13}, pages 446--461.
  Springer, 2014.

\bibitem{maji2013fine}
Subhransu Maji, Esa Rahtu, Juho Kannala, Matthew Blaschko, and Andrea Vedaldi.
\newblock Fine-grained visual classification of aircraft.
\newblock {\em arXiv preprint arXiv:1306.5151}, 2013.

\bibitem{xiao2010sun}
Jianxiong Xiao, James Hays, Krista~A Ehinger, Aude Oliva, and Antonio Torralba.
\newblock Sun database: Large-scale scene recognition from abbey to zoo.
\newblock In {\em 2010 IEEE computer society conference on computer vision and
  pattern recognition}, pages 3485--3492. IEEE, 2010.

\bibitem{cimpoi2014describing}
Mircea Cimpoi, Subhransu Maji, Iasonas Kokkinos, Sammy Mohamed, and Andrea
  Vedaldi.
\newblock Describing textures in the wild.
\newblock In {\em Proceedings of the IEEE conference on computer vision and
  pattern recognition}, pages 3606--3613, 2014.

\bibitem{helber2019eurosat}
Patrick Helber, Benjamin Bischke, Andreas Dengel, and Damian Borth.
\newblock Eurosat: A novel dataset and deep learning benchmark for land use and
  land cover classification.
\newblock {\em IEEE Journal of Selected Topics in Applied Earth Observations
  and Remote Sensing}, 12(7):2217--2226, 2019.

\bibitem{soomro2012ucf101}
Khurram Soomro, Amir~Roshan Zamir, and Mubarak Shah.
\newblock Ucf101: A dataset of 101 human actions classes from videos in the
  wild.
\newblock {\em arXiv preprint arXiv:1212.0402}, 2012.

\bibitem{loshchilov2016sgdr}
Ilya Loshchilov and Frank Hutter.
\newblock Sgdr: Stochastic gradient descent with warm restarts.
\newblock {\em arXiv preprint arXiv:1608.03983}, 2016.

\end{thebibliography}
\bibliographystyle{unsrt}

\end{document}